# SYLLABLE-BASED NEURAL NAMED ENTITY RECOGNITION FOR MYANMAR LANGUAGE

Hsu Myat Mo and Khin Mar Soe

Natural Language Processing Lab., University of Computer Studies,
Yangon, Myanmar

## ABSTRACT

*Named Entity Recognition (NER) for Myanmar Language is essential to Myanmar natural language processing research work. In this work, NER for Myanmar language is treated as a sequence tagging problem and the effectiveness of deep neural networks on NER for Myanmar language has been investigated. Experiments are performed by applying deep neural network architectures on syllable level Myanmar contexts. Very first manually annotated NER corpus for Myanmar language is also constructed and proposed. In developing our in-house NER corpus, sentences from online news website and also sentences supported from ALT-Parallel-Corpus are also used. This ALT corpus is one part of the Asian Language Treebank (ALT) project under ASEAN IVO. This paper contributes the first evaluation of neural network models on NER task for Myanmar language. The experimental results show that those neural sequence models can produce promising results compared to the baseline CRF model. Among those neural architectures, bidirectional LSTM network added CRF layer above gives the highest F-score value. This work also aims to discover the effectiveness of neural network approaches to Myanmar textual processing as well as to promote further researches on this understudied language.*

## 1. INTRODUCTION

Named Entity Recognition is the process of automatically tagging, identifying of labeling different named entities (NE) in text in accordance with the predefined sets of NE categories (e.g., person, organization, location and time and so on). NER has been a challenging problem in Myanmar language processing. Currently, Myanmar NLP is at an initial stage and lexical resources available are very limited. On the other hand, it has both complex and rich morphology as well as ambiguity. These facts can lead to wrong word segmentation. The current Myanmar-English Machine Translation system couldn't recognize names written in Myanmar language properly. Moreover, the most important motivation behind this work is that there is no currently available NER tool that can extract named entities in Myanmar scripts. As far as it is concerned, the previous proposed methods for Myanmar NER adopted dictionary look-up approach, rule-based techniques and a combination of rule-based and statistical N-grams based method together with name database.

The previous proposed methods for NER can be roughly classified into dictionary-based or rule-based and traditional statistical sequence labeling approaches and hybrid [1]. Those approaches require linguistic knowledge and feature engineering.

Neural network models have the advantage of minimizing the effort in feature engineering, since deep layers of neural networks can discover relevant features to tasks. General neural network architecture for sequence labelling tasks was developed and proposed in [2]. Following this work, Recurrent Neural Networks (RNN-based networks) have been designed for modeling sequential data and have been proved to be quite efficient in NER as a sequential tagging task. As a special kind of RNN, Long Short-Term Memory (LSTM) neural network has been verified to be powerful in modeling sequential data. Moreover, bidirectional LSTM neural network which is derived from LSTM network, introduces two independent layers to accumulate contextual information from the past and future histories. Bidirectional LSTM network has advantages in memorizing information for long periods in both directions so that it makes great improvement in linguistic computation. In [3], the authors combined a bidirectional LSTM network and a Conditional Random Fields (CRF) to form a BiLSTM_CRF network. This kind of model can use the past input features and sentence level tag information and the future input features. With the rapid development of deep learning, recent research revealed that those kinds of networks significantly outperform statistical algorithms.

In this work, NER is considered as a sequence labeling task and experiments are carried out on syllable level data. In order to explore the effectiveness of deep learning on Myanmar Language, experiments are conducted using bidirectional LSTM network by adding CRF inference layer above. We will refer that model as BiLSTM_CRF in the following sections. . In our experiments, we firstly encode syllable level information by using various neural networks such as Convolutional Neural Network (CNN) and bidirectional LSTM network on characters and then concatenate character and syllable level representation. The combined representation was fed into bidirectional LSTM network to learn context information of each syllable. On the top of bidirectional LSTM, a CRF inference layer was jointly added to decode the optimal label. Besides, we also performed the neural experiments with softmax layer on the top. That kind of model will be referred as BiLSTM in the following sections. Experiments are carried out with different parameters and network settings. Moreover, we tried to investigate which network (CNN and bidirectional LSTM) is more powerful in learning character representation. Experiments with these neural networks, without additional feature engineering, achieved promising results compared to the baseline CRF model. This study is the very first work to apply neural networks to Myanmar NER.

## 2. RELATED WORKS

As far as being aware and up to our knowledge, no work has been published for using deep neural networks on NER for Myanmar language. Previous effort on Myanmar NER had been done by rule-based and statistical research works. A method for Myanmar Named Entity Identification using a hybrid method was proposed in [4]. Their method was a combination of rule-based and statistical N-grams based method and name database was used as well. In [5], the authors proposed a Myanmar Named Identification algorithm that defines the names by using some of the POS information, NE identification rules and clues words in the left and/or the right contexts of NEs that carry information for NE identification. Their limitation is that input sentence must be with specified POS tags. As a weakness, sematic implication of proper names is defeasible. Moreover, those approaches require linguistic knowledge and feature engineering. CRF-based NER for Myanmar language can be seen in [6]. In this paper, CRF, one of the statistical approaches had been proposed for Myanmar NER task. Moreover, the authors tried to explore on various combination of features for this approach.

Although statistical approaches such as CRFs have been widely applied to NER tasks, those approaches heavily rely on feature engineering. The authors of [7] introduced two neural architectures for NER that use no features. One was based on bidirectional LSTMs and CRF, and another one was constructed using a transition-based approach. Likewise, a neural network architecture that automatically detects word- and character-level features using a hybrid bidirectional LSTM and CNN architecture to eliminate the need for most feature engineering was proposed in [8].

Over the past few years, various deep neural networks have been applied for NER on different languages, e.g., Italian neural NER [9], Mongolian neural NER [10], Japanese neural NER [11] and Russian NER [12] and so on. Likewise, deep neural architectures also have been applied on NER for different domains, e.g., Neural NER for Medical Entities in Twitter [13] and NER for Twitter Massages [14], Biomedical Neural NER [15] and disease NER [16]. Their works revealed that neural networks have the great capability for NER tasks and significantly outperforms statistical algorithms.

## 3. MYANMAR LANGUAGE

Myanmar language is the official language of the Republic of the Union of Myanmar and has more than one thousand years' history. According to the documents, the Myanmar script was descended from the Brahmi script of ancient South India. It belongs to the Sino-Tibetan language family. Myanmar scripts are written in sequence from left to right in which white space may sometimes be inserted between phrases but regular white space is not used between words or between syllables (see Figure. 1 for the example). Words in Myanmar language are composed of single or multiple syllables. Moreover, it can be said that a Myanmar syllable is a composition of multiple characters. The Myanmar scripts usually have 75 characters in total and those characters can be classified into 12 groups. For more details, you can check the paper [17].

Myanmar sentence: ရန်ကုန်တွင်မိုးမရွာပါ။
English sentence: It doesn't rain in Yangon.

Figure 1. Example of Myanmar writing

Words in Myanmar language are composition of one or more syllables and a syllable may also contain one or more characters. A word 'နိုင်ငံ' is composed of two syllables 'နိုင်' and 'ငံ'. A syllable in Myanmar language may be made up of one or several characters. For example, the syllable 'နိုင်' incudes five Unicode characters, i.e., 'န, ို, ်, င' and '်' (see Figure 2). In this paper, the character refers to the Unicode character instead of non-standard font characters.

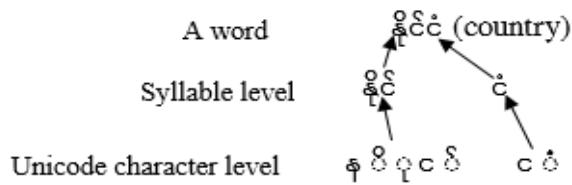

Figure 2. An example of a Myanmar word formation

For Myanmar language, there are some complex font problems. Most Myanmar people are more familiar with fonts that are not standard Unicode font. To make Myanmar syllable structure represented in a definite way, we have Unicode level as the character level in this work.

A Myanmar syllable structure formation is quite definite and simple. Although the constituents can appear in different sequences, a Myanmar syllable usually consists of one initial consonant followed by zero or more medials, zero or more vowels and optional dependent various signs.

Independent vowels, independent various signs and digits can act as stand-alone syllables. Moreover, only one consonant can even stand as a syllable. According to the Unicode standard, vowels are stored after the consonant. Detailed syllable segmentation structure for Myanmar language is explained in [17] and [18].

For example, the following Myanmar sentence, "ရန်ကုန်တွင်မိုးမရွာပါ။", contains seven syllables. The character "။" is sentence ending punctuation mark. White space is used between syllables (see Figure 3). The meaning of the sentence is "It doesn't rain in Yangon". "Yangon" is the city of the Republic of the Union of Myanmar. The proposed syllable-based sequential labeling model for Myanmar NER is trained and tested on the syllable level input data.

ရန် ကုန် တွင် မိုး မ ရွာ ပါ ။

Figure 3. Syllable-based segmented sentence

Myanmar language is very complex compared to English and other European languages. On the other hand, language resources for Myanmar NLP researches have not been well prepared until now. As one of agglutinative languages, Myanmar has complex morphological structures which can lead to suffer the models from data sparsity. Besides, for models in which words are treated as basic unit to construct distributed representation, there are problems for those rich morphological words. On the other hand, it is needed to deal with out-of-vocabulary words and word segmentation problem is one of the problems for Myanmar language. As described above, words in Myanmar language are not always separated by spaces, so word segmentation is necessary and segmentation errors will affect the level of NER performance. In Myanmar language, syllable is the smallest linguistic unit that can carry information about word. For those reasons, syllable is treated as the basic unit for label tagging in all our NER experiments.

### 3.1. CHALLENGES OF NER FOR MYANMAR LANGUAGE

The task of identifying names automatically in Myanmar text is complex compared to other languages for many reasons. One of the reasons is the lack of resources such as annotated corpus, name lists, gazetteers or name dictionary, etc. which means that Myanmar language is resource-constrained language. Besides, Myanmar language has distinct characteristics and having no capitalized latter which is the main indicator of proper names for some other languages like English. Moreover, its writing structure is of the free order, makes the NER a complex process. Some proper names of foreign person and location are loanwords or transliterated words so that there are wide variations in some Myanmar terms. Myanmar names also take all morphological inflections which can lead to ambiguity. This ambiguity of NE may lead to problem in classifying named entities into predefined types. It can be said that how to perform the task of identifying names in Myanmar text automatically is still challenging.

## 4. METHODOLOGY

### 4.1. CONVOLUTIONAL NEURAL NETWORK (CNN)

Convolutional Neural Network (CNN) is popular in dealing with images and it also has excellent results in computer vision. Lately, it has been applied in NLP research tasks and it can be found that it outperformed traditional models such as bag of words and n-grams and so on.

International Journal on Natural Language Computing (IJNLC) Vol.8, No.1, February 2019

### 4.2. RECURRENT NEURAL NETWORK (RNN)

Recurrent Neural Network (RNN) is powerful for learning sequential information. RNN does the same operation over and over again on stream of data. RNN comes with the assumption that the output is being dependent on the previous computations while a traditional neural network assumes all inputs (and outputs) are independent of each other. In theory, RNNs can make use of information in arbitrarily long sequences, but in practice they fail due to the gradient vanishing/exploding problems [19].

### 4.3. LONG SHORT-TERM MEMORY (LSTM)

LSTMs, a special kind of RNN, are designed to address the gradient vanishing problems of RNNs. Therefore, they are capable of learning long-term dependencies. They were introduced by [20]. The key to LSTM is the cell state which is controlled by three multiplicative gates. The formulas to update each gate and cell state of input $x$ are defined as follows:

$$i_t = \sigma(W_{xi} x_t + W_{hi} h_{t-1} + b_i) \qquad (1)$$

$$f_t = \sigma(W_{xf} x_t + W_{hf} h_{t-1} + b_f) \qquad (2)$$

$$o_t = \sigma(W_{xo} x_o + W_{ho} h_{t-1} + b_o) \qquad (3)$$

$$c_n = \tanh(W_{xc} x_t + W_{hc} h_{t-1} + b_c) \qquad (4)$$

$$c_t = f_t \odot c_{t-1} + i_t \odot c_n \qquad (5)$$

$$h_t = o_t \odot \tanh c_t \qquad (6)$$

where $\sigma$ denotes the sigmoid function; $\odot$ is the element-wise multiplication operator; $W$ terms denote weight metrics; $b$ are bias vectors; and $i, f, o$ denote input, forget and output gate respectively and $c$ are cell activation vectors. The term $h$ represents the hidden layer nodes.

### 4.4. BIDIRECTIONAL LSTM

BiLSTM network, a modification of the LSTM, is designed to capture information of sequential data and have access to both past and future contexts. It consists of two LSTMs, a forward and backward LSTM propagating in two directions. The forward LSTM reads stream of input data and computes the forward hidden states, while the backward LSTM reads the sequence in reverse order and creates the backward hidden states. The idea behind is that it creates two separate hidden states for each sequence so that the information of sequences from both directions is memorized. By concatenating the two hidden states, the final output is formed. This advantage of memorizing information for long periods in both directions can make great improvement in linguistic computation.

### 4.5. CONDITIONAL RANDOM FIELDS (CRF)

CRF is a statistical probabilistic model for structured prediction. It provides a probabilistic framework for labeling and segmenting sequential data, based on the conditional approach. Let $y$ be a tag label sequence and $x$ be an input sequence of words. Conditional models are used to label the observation input sequence $x$ by selecting the label sequence $y$ that maximizes the conditional probability $p(y|x)$. To do so, a conditional probability is computed:

$$p(y|x) = \frac{exp^{Score(x,y)}}{\sum_{y'} exp^{Score(x,y')}} \quad (7)$$

where *Score* is determined by defining some log potentials $\log \psi_i(x, y)$ such that:

$$Score(x, y) = \sum_i \log \psi_i(x, y) \quad (8)$$

In here, there are two kinds of potentials: emission and transition.

$$Score(x, y) = \sum_i \log \psi_{EMIT}(y_i \rightarrow x_i) + \log \psi_{TRANS}(y_{i-1} \rightarrow y_i) \quad (9)$$

In training, log probability of a correct tag sequence is maximized.

## 5. OUR WORK

Experiments are performed by comparing different neural models on syllable level text rather than word level and the performance results are compared with baseline statistical CRF model. In our neural training, we try to investigate various neural network architectures that automatically detect syllable and character level features using bidirectional LSTM, CNN and also GRU architecture to eliminate the need for most feature engineering. Both CNN and bidirectional LSTM networks have been investigated for modeling character level information.

### 5.1. TRAINING SETUP

Experiments were carried out as syllable-based sequence labeling, in which the LSTM-based neural networks were trained. In order to convert the NER problem into a sequence tagging problem, a label is assigned for each token (syllable) to indicate the named entities in training and test data. We used the Myanmar syllable segmentation algorithm "sylbreak" of [21] on sentences for the syllable data representation and syllable-based labeling.

### 5.2. DATA PREPARATION

As mentioned before, annotated corpus for Myanmar language are limited and scattered. For our experiment and further researches on Myanmar NER, we developed a manually annotated Myanmar NER corpus. There is no other available Myanmar NE corpus that has as much data as our NE corpus. For Corpus building, Myanmar news sentences were collected from online official websites and manually annotated according to defined NEs tags. Moreover, we also take Myanmar sentences from ALT-Parallel-Corpus because a lot of transliterated names appear in these sentences. Sentences from ALT corpus are translated from International news so that a lot of transliterated names are detected. The ALT corpus is one part of the Asian Language Treebank (ALT) project. Currently we have over 60K sentences in total. We separate those data into three sets, 58,043 sentences for training, 1,200 sentences for development and 1,200 sentences for testing. We defined six types of NE tags for manual annotation: PNAME, LOC, ORG, RACE, TIME and NUM.

PNAME tag is used to indicate person names including nickname or alias, while LOC tag is defined for location entities. In this case, location entities include politically or geographically defined places (cities, provinces, countries, international regions, bodies of water, mountain, etc.). Location also includes man-made structures like airports, highways, streets, factories and monuments. Names of organizations (government and non-government organizations, corporations, institutions, agencies, companies and other groups of people defined by an established organizational structure) are annotated with ORG tag. In our Myanmar language, some location names and names of national races have same spelling in writing scripts. For example, the location name "ကရင်" (Kayin State) and one of the national races "ကရင်" (Kayin race). For this reason, the NE tag RACE is defined to indicate names of national races. TIME is used for dates, months and years. NUM tag is used to indicate number format in sentences. Table 1 lists the entities distribution in our in-house NE corpus.

Table 1. Corpus Data Statistics.

| NE tags | Number of entities | | |
|---|---|---|---|
| | Train | Dev | Test |
| PNAME | 34262 | 622 | 517 |
| LOC | 60910 | 1365 | 1211 |
| ORG | 19084 | 375 | 281 |
| RACE | 5359 | 200 | 161 |
| TIME | 28385 | 556 | 530 |
| NUM | 19505 | 363 | 433 |

### 5.2.1. TAGGING SCHEME

In order to convert the NER problem into a sequence labeling problem, a label is assigned for each token (syllable) to indicate the NE in sentence. As tagging scheme, IOBES (Inside, Outside, Begin, End and Single) scheme is used for all the experiments.

### 5.3. EXPERIMENTS WITH CRF

For syllable-based CRF trainings, an open source toolkit for linear chain CRF [22] was used. Experiments were conducted by tuning various parameters with different features. Firstly, we only used the tokens and their neighbouring contents as features and set the window size as 5 and the best F-score of 89.05% is obtained when the cut-off threshold parameter is 3 and hyper-parameter c is 2.5. We also tried to add a small-sized named dictionary and clue words list as additional features into the experiment. The additional features make the F-score have around 2.4 % increase (91.47 %) compared to the F-score of 89.05 % (Table 2). It shows that CRF works the best when feature engineering is carefully prepared.

Table 2. Results on baseline CRF

| CRF models | Precision | Recall | F-measure |
|---|---|---|---|
| Syllable-based | 89.75 | 88.36 | 89.05 |
| Syllable-based+clue word feature+name list | 90.92 | 92.02 | 91.47 |

### 5.4. NEURAL MODEL TRAINING

Given a Myanmar sentence, syllable is treated as the basic training unit for label tagging. We first learn the input representation of each input token, and then the learned representation was fed into BiLSTM network for sequence leaning. On top of the network, a CRF inference layer determines the tag with the maximum probability of the syllable (see Figure. 4).

For the implementation of the neural network model training, we utilized the PyTorch framework [23] which provides flexible choices of feature inputs and output structure. Experiments were run on Tesla K80 GPU. Based on different parameter settings, the training time for each experiment is different.

As to input embedding setting, we tried experiments on using pre-trained 100-dim of embedding on syllable level and character level data, respectively. The data used for training the syllable and character embedding includes 200K sentences in total. In input embedding, there are two parts: character representation and syllable representation. Character embedding was tried to learn from training data by CNN and also with bidirectional LSTM network. Given a training sequence, we took syllables as basic training unit and syllables were projected into a d-dimension space and initialized as dense vectors.

As to optimization, both the stochastic gradient descent algorithm (SGD) and Adam algorithm were tried. For the SGD, it was performed with initial learning rate of 0.015 and momentum 0.1. The learning decay rate was set as 0.05. For the Adam algorithm, the initial learning rate was set as 0.0015. Both optimizations had batch sizes set as 30.

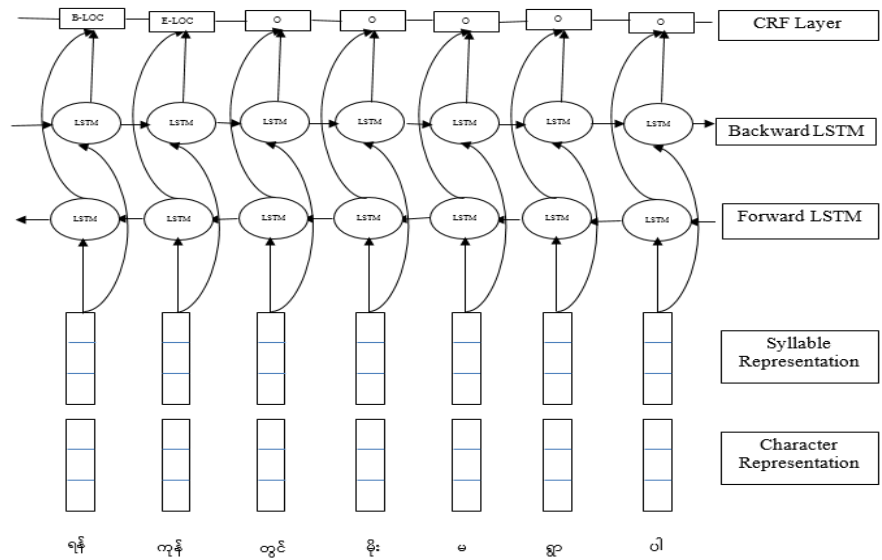

Figure 4. The architecture of our network

Early Stopping was used based on the performance on validation sets. A dropout of 0.5 was set for both embedding and output layers to mitigate overfitting in the training process. The hidden dimension was set to 200 in the whole experiment. For syllable-based BiLSTM_CRF which used CNN in character representation experiments, the best accuracy appears at 82 epochs and for the syllable-based BiLSTM_CRF which applied bidirectional LSTM on character representation experiments, the best accuracy happens at 77 epochs.

### 5.5. EXPERIMENTAL RESULTS AND ANALYSIS

In Table 3, we listed the best experimental results from different model structures. For both BiLSTM and BiLSTM_CRF on syllable level NER, we relied on a LSTM model or CNN model to learn char features first and then concatenated it with the syllables' embeddings as the input of to the BiLSTM or BiLSTM_CRF model. The char features learned from LSTM is not so good as CNN in our experiments. When the GRU is applied on syllable, it makes the F-score value drops continuously as the epoch increases, which is out of our expectation so we did not list the results in Table 3. The performance of using SGD optimization algorithm is slightly worse than using Adam, which you can see from Table 3. The random embedding also did not perform better than the pre-trained word embedding from word2vec. So the results in the Table 3 are from pre-trained embedding. For the syllable-based experiments, we can see that BiLSTM_CRF performs slightly better than BiLSTM.

As to the syllable based system, since syllables as inputs contain more information than individual characters as inputs, it is reasonable that it performs better than the character based system. On syllable level data, by using CNN to extract character features from the data as additional information inputs, better results are produced compared to not using character features or using LSTM to extract character features (the best performance is highlighted in Table 3).

If compared the results in Table 3 with the results in Table 2, we can see that neural models perform better than statistical CRF models. By comparison, syllable-based neural models without additional feature engineering perform better than CRF models. Syllable-based CRF model with additional feature is approaching the results of neural models. Although those experiments give the promising results, the size of data used for neural network model training is not so big compared to other NER corpus of other languages. Normally more data can help neural network learn better. Moreover, due to time and data limit, the hyper-parameters used in the experiment may be not the best. This needs modelling experiences and also large amount of trial and error experiments to decide.

Table 3. F-score results comparison between different neural models.

| Models | Precision/Recall/F-measure | |
|---|---|---|
| | Dev | Test |
| BiLSTM(SGD) | 89.40/88.48/88.94 | 89.99/89.97/89.98 |
| BiLSTM (Adam) | 89.32/88.47/88.89 | 90.71/91.64/91.17 |
| BiLSTM_CRF(Adam) with CharCNN | **91.06/90.45/90.75** | **94.66/94.99/94.82** |
| BiLSTM_CRF(Adam) with CharLSTM | 90.17/89.68/89.92 | 94.71/94.83/94.77 |
| BiLSTM_CRF(SGD) with CharCNN | 89.57/90.37/89.97 | 93.01/94.19/93.61 |
| BiLSTM_CRF(SGD) with CharLSTM | 84.82/85.65/85.65 | 90.24/91.83/91.03 |

## 6. CONCLUSION

In this paper, we have explored the effectiveness of neural network on Myanmar NER and conducted a systematic comparison between neural approaches and traditional CRF approaches on our manually annotated NE corpus. Experiment results revealed that the performance of neural networks on Myanmar NER is quite promising, because neural models did not use any handcrafted features or additional resources. Although our NE corpus is not so big, neural network models produce better performance than CRF models for Myanmar NER, we still believe with more data and more experiments, neural networks can learn better so as to produce better results. From the experiments, we can see that neural network performs much better while in the experiments of using CRF models, only by adding additional name list features and clue word list, produced the similar accuracy as the syllable-based neural models.

Anyway, this exploration of using neural networks for Myanmar NER is the first work to apply neural networks on Myanmar language. It showed us that BiLSTM_CRF network on syllable level data jointly with CNN to extract character feature can facilitate Myanmar NER. With more data and more experiments, better results will be reported in the future and we will keep exploring neural networks on other Myanmar NLP work, e.g., POS tagging and word segmentation, too.

**ACKNOWLEDGEMENTS**

The ASEAN IVO (http://www.nict.go.jp/en/asean_ivo/index.html) project "Open Collaboration for Developing and Using Asian Language Treebank" was involved in the production of the contents of this publication and financially supported by NICT (http://www.nict.go.jp/en/index.html).

**AUTHORS**

Hsu Myat Mo got her B.C.Sc (Hons) in 2010, followed by M.C.Sc (Credit:) in 2012, respectively. Currently she is doing her Ph.D research focusing on Myanmar NER in Natural Language Processing Lab, at the University of Computer Studies, Yangon.

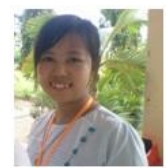

Dr. Khin Mar Soe got Ph.D(IT) in 2005. Currently she is working as a Professor and also the Head of Natural Language Processing Lab, at the University of Computer Studies, Yangon. She has been supervising Master thesis and Ph.D researches on Natural Language Processing. Moreover, she participated in the project of ASEAN MT, the machine translation project for South East Asian languages.

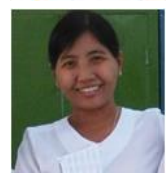